\documentclass[10pt,twocolumn,letterpaper]{article}

\usepackage{iccv}
\usepackage{times}
\usepackage{epsfig}
\usepackage{graphicx}
\usepackage{amsmath}
\usepackage{amssymb}
\usepackage{comment}
\usepackage{ulem}

\iccvfinalcopy 

\def\iccvPaperID{14} 

\ificcvfinal\fi
\begin{document}

\title{Road Scene Understanding by Occupancy Grid Learning\\ from Sparse Radar Clusters using Semantic Segmentation }
\author{Liat Sless \hspace{2.5cm}Gilad Cohen \hspace{2.5cm}Bat El Shlomo \hspace{2.5cm}Shaul Oron\\
General Motors, Advanced Technical Center, Israel\\
{\tt\small \{liat.sless, gilad.cohen, batel.shlomo, shaul.oron\}@gm.com}\\
}

\maketitle
\begin{abstract}

Occupancy grid mapping is an important component in road scene understanding for autonomous driving. It encapsulates information of the drivable area, road obstacles and enables safe autonomous driving.
Radars are an emerging sensor in autonomous vehicle vision, becoming more widely used due to their long range sensing, low cost, and robustness to severe weather conditions.
Despite recent advances in deep learning technology, occupancy grid mapping from radar data is still mostly done using classical filtering approaches.
In this work, we propose learning the inverse sensor model used for occupancy grid mapping from clustered radar data. This is done in a data driven approach that leverages computer vision techniques. This task is very challenging due to data sparsity and noise characteristics of the radar sensor. The problem is formulated as a semantic segmentation task and we show how it can be learned using lidar data for generating ground truth.
We show both qualitatively and quantitatively that our learned occupancy net outperforms classic methods by a large margin using the recently released NuScenes real-world driving data.

\end{abstract}

\begin{figure}[ht]
    \centering
    \begin{tabular}{c c c c}
         \includegraphics[width=1.7cm, height=6.2cm ]{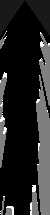} & 
         \includegraphics[width=1.7cm, height=6.2cm ]{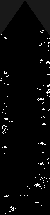} &
         \includegraphics[width=1.7cm, height=6.2cm ]{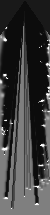} &
         \includegraphics[width=1.7cm, height=6.2cm ]{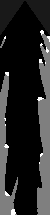} \\
         (a) & (b) & (c) & (d) \\
    \end{tabular}
    \caption{Occupancy grid learning. (a) Ground truth. (b) Aggregated radar input. (c) Classic ISM result (d) Occupancy net output (ours). The occupancy net learns a complex inverse sensor model (ISM) function that can handle sparse and noisy radar measurements. The network significantly outperforms the commonly used occupancy grid mapping algorithm using Bayesian filtering and hand crafted ISM function. Note: White is occupied, black is free, light gray is unobserved, dark gray is ignore.}
    \label{fig:teaser}
\end{figure}

\section{Introduction}\label{sec:introduction}
 Inferring the drivable area in a scene is a fundamental part of road scene understanding. This task is sometimes also referred to as "general obstacle detection", "free space estimation" or "occupancy grid mapping" \cite{elfes1989using}.
 
Autonomous vehicles are required to operate in complex environments. To this end, occupancy grid mapping from \textit{various} sensors is an important component of autonomous vehicle perception \cite{werber2015automotive},  \cite{lu2019monocular}, \cite{lombacher2017semantic}, \cite{li2018high}, \cite{dickmann2016automotive}, \cite{weston2018probably}. 
In real world scenarios, training class specific detectors for all possible object classes a vehicle might encounter is intractable. This in turn makes occupancy grid mapping a key capability, crucial for ensuring the vehicle avoids obstacles of unknown classes such as trash piles, boxes and other random obstacles.

Free space detection can be performed using data from different sensors. Lidars are especially suitable for this task due to their dense and accurate range measurements \cite{weiss2007robust}, and are widely used for autonomous driving applications. In recent years Radars are starting to emerge and become more frequently used in autonomous driving in general, and specifically for occupancy grid mapping. This is mostly due to their long range, low price, and performance under severe weather conditions such as fog, rain and snow \cite{dickmann2016automotive}, \cite{werber2015automotive}.

Traditionally, occupancy grid mapping is performed by applying Bayesian filtering and using hand-crafted inverse sensor model (ISM) functions \cite{thrun2005probabilistic}. In recent years, advances in deep learning have significantly improved the performance of machine vision applications. Despite this progress, deep learning in occupancy grid mapping is still not widely used. Moreover, when considering occupancy grid learning from radar data almost no prior work is to be found.

Typically, automotive radars produce sparse detections, in one of three forms: raw, clustered and object level data.
Raw level is the richest form of the data but is also the nosiest. 
Applying algorithms such as DBSCAN as in \cite{kellner2012grid}, or CFAR (constant false alarm rate) \cite{skolnik2008radar} on raw radar data results in clustered radar data, which is sparser but less noisy.
Last, object level data is the most processed, in which the detections are filtered and associated both in space and in time.
Generally, radar detections are statistical in nature and suffer from various noise factors.

In this work we focus on the challenging task of learning occupancy grid mapping from \textit{clustered} radar data. With many low cost commercial automotive radars providing only clustered data with no elevation information, this task is of high interest for autonomous driving. 

Inspired by advances in computer vision, we propose learning occupancy grid mapping for static obstacles, from radar cluster data, in a supervised manner. We formulate the problem as a semantic segmentation task with three classes: occupied, free and unobserved. This formulation explicitly handles regions in space for which we do not have measurement information, which is an inherent problem in this setting. Using this formulation, we learn an inverse sensor model function from data. Our method takes advantage of the spatial and temporal dependencies in the data in order to overcome data sparsity and noise.

We treat the radar as a 3D imaging sensor. To this end, we provide the data to the network as a birds-eye-view (BEV) image with desired grid resolution. Temporal information is leveraged by means of radar frame aggregation. The network is trained to infer the output occupancy grid mapping, learning the inverse sensor model function. Training labels are generated using lidar data.

An additional challenge is the problem of significant class imbalance, as free space is much more frequent than occupied space. We handle this problem by using a recently proposed surrogate of the intersection-over-union (IoU) loss called the Lovasz loss \cite{berman2018}. 

We demonstrate the efficacy of the proposed approach using the NuScenes data set \cite{caesar2019nuscenes} comprised of challenging real world driving scenes. We train the proposed occupancy grid network and demonstrate its performance both qualitatively and quantitatively. We show significant gains compared with standard filtering based methods for occupancy grid mapping using hand crafted ISM functions. 

To summarize, the main contribution of this work is:
(i) A method of processing sparse clustered radar data using computer vision techniques, used for road scene understanding by means of learning occupancy grid mapping that outperforms standard methods. This method learns an ISM function from data using multi frame aggregation to handle sparsity and noise. (ii) Proposing a method for performing supervised learning using lidar data to generate training labels while resolving issues related to range and coverage differences between lidar and radar. (iii) Quantitative results showing the efficacy of our approach for the occupancy grid mapping task on the recently released NuScenes data set.

\section{Related Work}\label{sec:related work}

Traditionally,   occupancy  grid  mapping  is  performed using  an  inverse  sensor  model  (ISM)  and  by  applying Bayesian  filtering techniques. Among the classical methods, a delta function ISM is typically used for lidar occupancy grid generation \cite{garcia2008high}, \cite{thrun2005probabilistic}, \cite{rexin2017modeling}, \cite{hoermann2018dynamic}. Since radar data is noisier, it is common to see a Gaussian variant (in range and azimuth) of the delta function ISM \cite{garcia2008high}, \cite{prophet2018adaptions}, \cite{werber2015automotive}. 
Algorithms operating on raw radar data are often required to clear noise and clutter, assess the detection probability / plausibility \cite{li2018high}, \cite{werber2015automotive}, and perform clustering on the results before generating the occupancy grid mapping \cite{li2018high}.

To the best of our knowledge, this work is the first to utilize clustered radar data using a learnable model, which is an important use case for autonomous driving. Other works concentrate on raw radar data \cite{garcia2008high}, \cite{prophet2018adaptions}, \cite{werber2015automotive},\cite{weston2018probably}, \cite{dickmann2016automotive} or object level data for high level fusion \cite{gohring2011radar}, \cite{chavez2016multiple}. 

In most, if not all, related work it is common to see a use of non-public data sets. Moreover, due to the lack of data, 
occupancy grid results are often demonstrated only qualitatively on few samples \cite{werber2015automotive},\cite{li2018high}, \cite{rexin2017modeling},  making it more difficult to compare between different published results. Other studies suggest measuring distance errors for specific scenarios \cite{degerman20163d}, or ROC curve \cite{hoermann2018dynamic}. Using mean IoU, of occupied and free regions, as a metrics for radar occupancy grid is done in \cite{weston2018probably}. Using it for generating occupancy grids from camera and lidar is proposed in \cite{oh2016fast}, \cite{lu2019monocular}.

In addition to identifying occupied and free cells, the inability to observe a known state of cells is also an important concept in occupancy grid mapping and has been addressed in several ways. In cases where each cell is associated with an occupancy probability, such as in \cite{werber2015automotive}, \cite{dickmann2016automotive}, \cite{prophet2018adaptions}, a probability of $0.5$ represents the highest uncertainty between occupied and free, and is equivalent to having no knowledge of a cell's occupancy state. 
Unobserved state can also be leveraged in inverse sensor modeling \cite{rexin2017modeling}.
In contrast, we define a class for unobversed cells, as a part of a semantic segmentation problem formulation.

\begin{figure*}[t]
    \centering
    \begin{tabular}{p{0.48\textwidth} p{0.48\textwidth}}
         \includegraphics[width=0.45\textwidth]{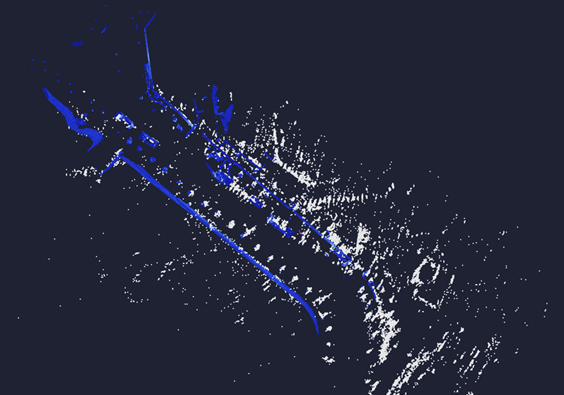} &  
         \includegraphics[width=0.45\textwidth]{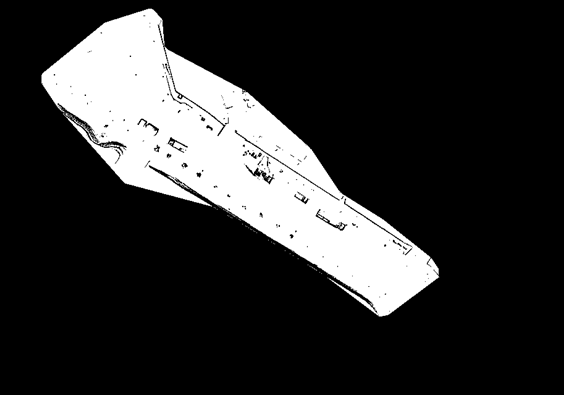} \\
         \centering (a) lidar and radar aggregated data.  &
         \centering (b) lidar point cloud concave hull.\\
    \end{tabular}
    \caption{Real world scene from the NuScenes data set. (a) Showing an overlay of aggregated lidar (Blue) and Radar (white) data. Radar data is much noisier than lidar data. Also, due to the longer range of the radar there are a lot of radar points in regions which are beyond what the lidar can observe. (b) Concave hull used to mask regions without lidar data in training and evaluation. See text for more details.}
    \label{fig:data} 
\end{figure*}

\section{Occupancy grid mapping}\label{sec:method}
Occupancy grid mapping is an important part of road scene understanding and is used for generating consistent estimation of free and occupied space from noisy sensor measurements. The measurements themselves can come from different sensors, providing information on the state of the world as well as the vehicle pose, using inertial measurement units (IMU) and vehicle odometry.

Occupancy grid maps partition the space in front or around the vehicle into a fine grained grid. Each grid cell is represented by a random variable corresponding to the occupancy of the location it covers. The grid map is usually represented in the relevant sensor's coordinate frame, or in the host vehicle's coordinate frame where sensor fusion is desired \cite{thrun2005probabilistic}. 

Let $y \in \{occupied, free, unobserved\}^{H \times W}$ be a grid map of size $H \times W$, with spatial resolutions $\alpha_x, \alpha_y$. We denote by $y^{u,v}$ the occupancy state of each cell $(u,v)$. 

Occupancy grid mapping algorithms aim to calculate the posterior probability of the grid map $y$ given the measurement data:
\begin{equation}\label{eq:og1}
	p(y|z_{1:t},x_{1:t})
\end{equation}
Where $z_{1:t}$ is the set of all measurements up to time $t$, and $x_{1:t}$ is the sequence of host vehicle poses.

In this work we focus on the task of performing occupancy grid mapping for static obstacles from clustered radar data. This type of data is typically very sparse and does not provide elevation information. This is a challenging setup which is notably relevant to the use of automotive radars. This problem relates to the most common type of occupancy grid maps, that of 2-D planar maps viewing space from a birds-eye-view (BEV). 

\subsection{Bayesian filtering and inverse sensor model}\label{sec:ism}
We now provide a short introduction to the canonical occupancy grid mapping approach, and refer the reader to \cite{thrun2005probabilistic} for a more thorough derivation.

The classic occupancy grid mapping algorithm has two main elements. The first is an ISM function dictating how a given measurement affects the occupancy state. The second is Bayesian filtering which governs how cell occupancy is updated over multiple temporal samples.

Estimating the probability in equation \eqref{eq:og1} is difficult. In order to make this problem tractable it is first assumed that the state of each cell is independent with respect to other cells. This allows breaking down the problem of estimating the map into a collection of separate problems such that:
\begin{equation}
	p(y|z_{1:t},x_{1:t}) = \prod_{u,v}p(y^{u,v}|z_{1:t},x_{1:t})
\end{equation}
Instead of working directly with probability values it is common to use the log-odds representation of occupancy:
\begin{equation}
	l_t(y^{u,v}|z_{1:t},x_{1:t}) = log(\frac{p(y^{u,v}|z_{1:t},x_{1:t})}{1-p(y^{u,v}|z_{1:t},x_{1:t})})
\end{equation}
Using Bayesian filtering one can obtain a formulation for updating the occupancy probability of a cell $(u,v)$ over time, using log-odds, as follows:
\begin{multline}
l_t(y^{u,v}|z_{1:t},x_{1:t}) = l_t(y^{u,v}|z_{t},x_{t})+\\
l_{t-1}(y^{u,v}|z_{1:t-1},x_{1:t-1})-l_0
\end{multline}
Where the term $l_t(y^{u,v}|z_{t},x_{t})$ represents the inverse sensor model defining how grid cells are updated given observations, and the constant $l_0=log\frac{p(y^{u,v}=1)}{p(y^{u,v}=0)}$ is the occupancy log-odds prior.

The most common ISM functions used for lidar and radar data include the Delta and Gaussian functions. Using a Delta function, for example, means that if a return was obtained from some cell $(u,v)$, then that cell is updated with a higher probability of being occupied. Cells along the line-of-sight between the sensor and $(u,v)$ are updated with a higher probability of being free. Finally, cells along the line-of-sight after $(u,v)$ are considered unobserved and therefore left unchanged.

\subsection{Occupancy grid learning}\label{sec:og_learning}
Instead of using hand crafted ISM functions (like in the classical method), we propose learning an inverse sensor model function in a data driven manner using deep learning. Additionally, since our data is extremely sparse, we propose learning this ISM over an aggregation of multiple radar frames. Specifically, this means our occupancy network learns to infer: 
\begin{equation}
	p(y|z_{t-k:t},x_{t-k:t},\theta)
\end{equation}
Where $\theta$ are neural network parameters optimized in a supervised learning process, and $k$ the number of aggregated radar frames. 

The occupancy net learns the underlying joint probability function from data. Note that this is unlike the classical approach where we need to assume grid cells are independent.  

Another important observation is that temporal information is directly incorporated into the ISM function itself (and not by means of Bayesian filtering). To use multiple frames, we compensate for the ego motion between frames using the pose information $x_{t-k:t}$, warping all frames to the coordinate frame of the host vehicle at time $t$. 

When using an occupancy grid mapping algorithm one eventually obtain the posterior probability for each cell which resides in the range $[0,1]$. This real valued output captures not only if a cell is likely to be occupied (near 1) or free (near 0), but also if the cell is actually observable by the sensor. When a certain cell is unobserved the resulting probability will be close to 0.5 meaning we do not know whether it is occupied or free. This is reflected in the classical algorithm by the fact cells are only updated if they are observed, and also by the occupancy prior.

In order to explicitly incorporate this observability information into our model we formulate the learning problem as a 3 class semantic segmentation task. Specifically, for each grid cell the network predicts one of 3 possible labels: occupied, free or unobserved which corresponds to occupancy probabilities of 1.0, 0.0 and 0.5 accordingly.

Another major challenge inherent to occupancy grid mapping is significant class imbalance. Namely, there is a much larger a-priory probability of seeing free space relative to occupied space. This in turn means there are significantly less grid cells with the occupied label compared to those with free or unobserved labels. 

We propose using a metric that explicitly addresses the problem of class imbalance. This metric is the intersection-over-union (mIoU) per class, which is commonly used in semantic segmentation problems  \cite{ronneberger2015u}, \cite{weston2018probably}:

\begin{equation}\label{eq:miou_loss}
    IoU_c = \frac{|\{y=c\}\cap\{ \tilde{y}=c\}|}{|\{y=c\}\cup\{\tilde{y}=c\}|}  
\end{equation}
Where $y$ is the predicted occupancy map,$\tilde{y}$ are the true labels and $c\in C, C=\{occupied, free, unobserved\}$ is the class. In this case an overall metric giving equal weight to all classes will be: 
\begin{equation}\label{eq:final_loss}
    mIoU = \frac{1}{|C|}\sum_{c} IoU_{c} 
\end{equation}

The problem of class imbalance relates to both performance evaluation as well as to learning. If class imbalance is not addressed as part of a network's training loss function, the network might converge to a solution which never predicts occupied cells. One example is the common cross entropy loss, where this issue is usually addressed by some form of weighting, which involves adding hyper parameters that are not necessarily easy to tune.    

We would like our network to obtain results minimizing the above metric. Unfortunately, directly minimizing over the IoU complement which is the complement of equation \eqref{eq:miou_loss} is not possible since it is not differentiable. Instead we use the recently proposed Lovasz loss \cite{berman2018} which is a surrogate to the intersection-over-union loss. It extends the notion of IoU from discrete to continues space making it differentiable, and therefore suitable for back-propagation used in stochastic gradient decent. Lovasz loss was shown in the original paper to be useful for semantic segmentation learning. Using it as proposed in equation \eqref{eq:final_loss} automatically scales all classes to have equal weight regardless of the a-priory data distribution. This in turn helps the training to optimize for all classes without adding any additional hyper parameters.  
\begin{figure}[t]
    \centering
    \includegraphics[width=0.48\textwidth]{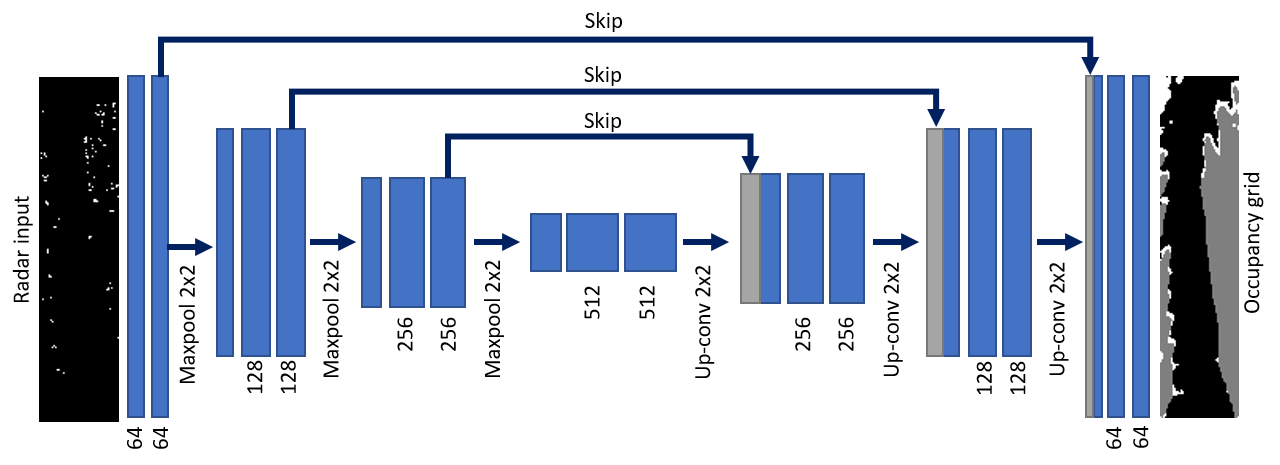}
    \caption{Network architecture. We propose an encoder-decoder architecture with skip connections to help maintain fine grained information. The network input are radar clusters embedded as a BEV image and its output is the corresponding occupancy grid map (with the same resolution). Note: Numbers correspond to the number of filters in each layer. All convolutions are 3x3.}
    \label{fig:unet}
\end{figure}
\section{Experiments}\label{sec:experiments}
We begin by describing the procedure of generating training labels using lidar data in section \ref{sec:Labeling_procedure}. We then discuss the experimental setup in section \ref{sec:Experimental_setup} including model architecture, data, training details and performance evaluation. Finally, results are shown in section \ref{sec:results}.

\subsection{Labeling procedure} \label{sec:Labeling_procedure}
In this section we describe an automated labeling process. Its goal is generating ground truth labels for the occupancy of each cell in the grid map by using accurate lidar data \footnote{ \url{https://github.com/liat-s/radar_occupancy_grid/}}. As part of this procedure we also generate a mask which solves the problem of handling regions of space for which we have radar data but do not have lidar data. 

For each radar frame, we generate a grid map representation containing the true state of occupancy for each cell,  $\tilde{y} \in \{occupied, free, unobserved\}^{H \times W}$. Each such grid map is represented in the matching radar's coordinate frame. 

Lidar range is typically limited relative to radar. Therefore, we aggregate over all lidar frames in a given scene. This enables the ground truth data not to be strictly limited by the range of the lidar in a specific frame, and also helps overcome temporary occlusions.  

The aggregated point cloud is represented in global coordinates.
For each radar frame, we transform the point cloud to the radar's coordinate frame. We then project the aggregated $3D$ point cloud onto a $2D$  $H \times W$ grid, with spatial resolutions $\alpha_x, \alpha_y$. Noise and dynamic obstacles are filtered out using binary thresholding clearing cells with low point count. Next, we apply morphological operations for additional noise removal and smoothing. Specifically, we perform: dilation, hole filling, and erosion. This results in a grid representation containing aggregated information for all static obstacles in the scene. 

Given the positions of all static obstacles on the grid, we map the occupancy state of each grid cell from a specific radar viewpoint using ray tracing. Specifically, we consider all cells between the radar and the first return along a ray as free. All consecutive occupied cells along the ray are considered as a single obstacle and label as occupied. Cells after the first obstacle are marked as unobserved. 
This approach is similar to that taken in \cite{weston2018probably}. The main difference is we do not distinguish between partially observed and unobserved cells, considering anything after the first return as unobserved. Finally, areas of the occupancy map that cannot be observed by the radar, due to its limited field-of-view angle, are masked out indicating they will not be used for loss computation or metric evaluation.

The last issue we address when working with real world recordings, is that lidar and radar coverage is different along the scene boundaries. This is mainly due to range limitations resulting in some parts of space having radar data, but not lidar data. This phenomenon is evident in directions perpendicular to the driving direction and near the end of the recording.

In order to use grid maps which contain partial lidar ground truth, we propose computing a concave hull \cite{edelsbrunner1983shape} (also known as a concave closure or an alpha shape) over the projected lidar point cloud. We then only label cells that are contained within the concave hull. The rest of the cells (outside the concave hull) are marked with an ignore value meaning they will not be used for loss or metric computations. Figure \ref{fig:data} (a) shows a typical scene. There are many regions where only radar data is available (no lidar returns). Computing loss or metrics in these regions will results in erroneous training and evaluation. The corresponding concave hull used for masking this scene is shown in figure \ref{fig:data} (b). Notice how this hull tightly bounds the region of the scene covered by lidar data. 

\subsection{Experiment setup}\label{sec:Experimental_setup}
\begin{figure}
    \centering
    \includegraphics[width=0.48\textwidth]{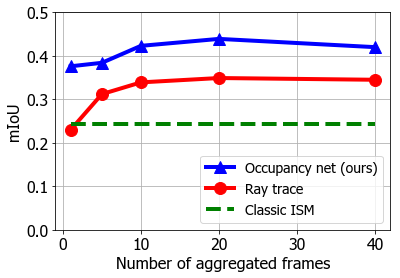}
    \caption{Performance for different numbers of aggregated radar frames. Occupancy net (Blue curve) significantly outperforms ray tracing (Red curve), using the same data, as well as the best result obtained by the classic approach using ISM (green line). In fact, the network using just a single radar frame outperforms ray tracing with \textit{any} number of aggregated frames. This indicates our occupancy net is learning an ISM function that is more complex than simple ray trace, helping it overcome radar noise and data sparsity.}
    \label{fig:radar_agg}
\end{figure} 
\newcommand{\partitle}[1]{\noindent\textbf{#1}}
\partitle{Model architecture:} Inspired by U-net \cite{ronneberger2015u} we propose using an encoder-decoder architecture with skip connections as shown in figure \ref{fig:unet}. This type of network was previously shown to be effective for learning semantic segmentation tasks \cite{lin2017feature, badrinarayanan2017segnet}.
\medskip

\partitle{Data:} We demonstrate the performance of our approach using the NuScenes data set \cite{caesar2019nuscenes} comprised of real world driving scenes. Each scene containing 20 seconds of recorded data from 6 cameras, 1 lidar and 5 radars providing clustered data output. Each radar outputs up to 128 clusters providing their 2D location (no elevation) and velocity. The velocity is used to filter out dynamic clusters. In total, we use 100 scenes removing 4 scenarios where the host vehicle is static or our view is blocked for the entire scene. The data is split into training ($80\%$), validation ($5\%$) and test sets ($15\%$). 
\medskip

\partitle{Training:} We train the occupancy net to output a grid of size $H\times W=215\times50$ in the radar coordinate frame. For our experiments we use grid cell size $\alpha_x=\alpha_y=40cm$. The resulting occupancy grid map covers an area from 0m to 86m in range, and 20m in width ranging from -10m to 10m. The network input is a grid of similar size where each grid cell has a binary value and is equal to one if a radar cluster resides in that cell and zero otherwise. As previously mentioned we aggregate radar data using $k$ consecutive frames $z_{t-k:t}$. In order to compensate for the host's ego motion, each frame is warped to the last ego pose $x_t$ before performing the aggregation. The radar frames are highly correlated temporally, therefore, training examples are taken such that they are in non-overlapping windows, i.e. $\{z_{1:k}, z_{k+1:2k},...\}$.

Our network model is trained using the Lovasz loss as explained in section \ref{sec:og_learning}. In addition, we ran ablation tests using the cross entropy loss, with varying weights to handle class imbalance.

Data from all 5 radars is used meaning the network is agnostic to radar mounting. Thus the same network can be used to infer the occupancy grid map of any of the 5 radars.  

Ground truth labels are generated as explained in section \ref{sec:Labeling_procedure} using aggregated lidar data. The network is trained using SGD with momentum optimizer \cite{qian1999momentum}. We set the initial learning rate to $0.05$ and momentum $0.9$. The learning rate was decayed by a factor of 0.9 wherever the mean intersection-over-union (mIoU) metric plateaued for two epoches. Horizontal flip data augmentation is used to increase the data set size.
\medskip

\partitle{Baseline methods:} We compare the performance of our occupancy grid mapping network with the commonly used classic approach that uses Bayesian filtering and an ISM as described in section \ref{sec:ism}. We use two ISM functions which are commonly used in the literature. The first is the Delta function ISM and the second in the Gaussian ISM. For each method we performed hyper parameter search using the validation set. The aim is to find the best thresholds used to determine which cells are occupied, free and unobserved based on the cell's posterior probability estimate.  

In order to emphasize that our network is learning a complex ISM function, we provide an additional ray tracing baseline. Specifically, we take the aggregated radar data, used as network input, and instead of feeding it to the network we perform ray tracing in a similar manner to what is described in section \ref{sec:Labeling_procedure}.  
\medskip

\partitle{Evaluation metric:} We qualitatively compare between the different methods using the IoU metric as shown in equation \eqref{eq:miou_loss}. The IoU is computed per each of the 3 classes (free, occupied and unobserved). Then the mIoU value over the 3 classes is also computed as shown in equation \eqref{eq:final_loss}. This provides one figure of merit for overall performance.

\subsection{Results}\label{sec:results}
Results comparing the performance of all method on the test set data are presented in table \ref{tab:results}. As can be seen our occupancy net outperforms the classic approaches using Bayesian filtering and ISM models by a large margin. The limited performance of the classic methods in this setup demonstrates the difficulty of performing occupancy grid mapping with such sparse and noisy radar data. The network also significantly outperforms direct ray tracing on the input data in each of the classes resulting in over $25\%$ improvement in mIoU. 

We note that ray tracing outperforms classic ISM approaches and hypothesize there are two main reasons for this. The first is data sparsity. Unlike the ray tracing which uses aggregated radar data, which reduces sparsity, the classic ISM algorithms works frame by frame, seeing very sparse data at each stage. Supporting this hypothesis is the observation that ray tracing with just a single frame produces similar results to classic ISM, as can be seen in figure \ref{fig:radar_agg}. The second reason ray tracing outperforms the classic ISM is related to differences in their underlying assumption regarding the nature of the world. Specifically, the classic ISM approach uses an a-priori assumption that the world is unobserved. This means that cells are considered unobserved until evidence is provided. In contrast the ray tracing algorithm uses an a-priori assumption that the world is free. A ray along which there are no returns (no radar points) will be considered all unobserved by the classic ISM while considered all free by the ray tracing. When the data is very sparse, as in our case, this difference in a-priori assumption, dramatically affects the results. An example of this phenomenon can be seen in figure \ref{fig:teaser}.   

One of the main factors impacting the performance of the occupancy net is the number of aggregated radar frames used as input. As can be seen in figure \ref{fig:radar_agg} (Blue curve), using 10 or 20 aggregated radar frames provides a $12.5\%$ and $16.7\%$ improvement in mIoU, over using a single radar frame, accordingly. Best results are obtained for 20 frames after which point results start to drop.

\begin{figure}[h]
     \centering
     \begin{tabular}{c c c c}
          \includegraphics[width=1.7cm, height=6.2cm ]{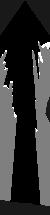} & 
          \includegraphics[width=1.7cm, height=6.2cm ]{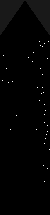} &
          \includegraphics[width=1.7cm, height=6.2cm ]{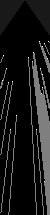} &
          \includegraphics[width=1.7cm, height=6.2cm ]{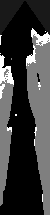} \\
          (a) & (b) & (c) & (d) \\
     \end{tabular}
     \caption{Learning with single frame input. (a) Ground truth. (b) Radar input (single radar frame). (c) Ray trace result (d) Occupancy net output. The occupancy net learns a complex ISM function able to infer the occupancy state of the world from very sparse data. This is evident when comparing the network results to simple ray tracing performed on the same input. Note: White is occupied, black is free, light gray is unobserved, dark gray is ignore.}
     \label{fig:one_frm}
\end{figure}

We point out that the occupancy network consistently outperforms ray tracing on the aggregated input. This shows that our network is learning an intricate inverse sensor model function that is much more complex then simple ray tracing. It is this ISM learning that allows the network to overcome data sparsity and radar noise. Even using just a single radar frame is enough for the occupancy network to outperform ray tracing with \textit{any} number of aggregated frames, as can be seen in figure \ref{fig:radar_agg}. 
A qualitative example demonstrating the kind of ISM function the network is able to learn using just a single radar frame is presented in figure \ref{fig:one_frm}. When compared to simple ray tracing, with the same input, it is evident that the network learns a complex spatially meaningful ISM. It can be seen that the network learns to ignore points in unobserved space while expanding other points to generate clear and continuous occupied regions.     
With regards to the network loss used, after fine tuning the weights of the cross entropy loss, we achieved similar results (less than $0.001$ difference in metrics). Since using Lovasz does not require fine tuning to handle class imbalance, it is less time consuming, more robust and is our recommended setting.    

Qualitative results are shown in figure \ref{fig:results}. It can be seen that the radar input is very noisy (second row) even when aggregating over 20 frames. Notice how our occupancy net (last row) is able to produce much sharper results than those of the classic ISM approach (third row). In some cases the classic ISM produces clutter in the form of isolated occupied points in free regions. This seems to be related to radar clusters which are generated due to reflections from the road. This noise is hard to filter without elevation information. The occupancy net, leveraging the labels produced by lidar, does not suffer from this problem, producing much cleaner output.   
We note that in some cases the network generates undesired blobs of different classes. For example the free region at the bottom-left of subplot (f) and free blobs in subplot (d). Small unobserved blobs in the free area of subplots (b) and (c) are another example. We believe this behavior is caused by the network predicting each pixel class separately without any explicit spatial consistency mechanism. In future work we plan to try and fix this behaviour by adding additional loss terms.   

\begin{table}[h]
    \centering
    \begin{tabular}{l|c|c|c|c}
         Method         & Occupied & Free  & Unobs.  & 
         mIoU\\\hline
         Delta ISM      & 0.029    & 0.391 & 0.311   &  0.244  \\
         Gaussian ISM   & 0.012    & 0.444 & 0.213   &  0.223  \\
         Ray trace      & 0.066    & 0.576 & 0.405   &  0.349  \\
         Occupancy net & \textbf{0.108}  & \textbf{0.614} & \textbf{0.593}   &  \textbf{0.439}
    \end{tabular}
    \caption{Quantitative results on the test set. Comparing our occupancy net (last row) to classic Bayesian filtering approaches using ISM and to performing ray tracing directly on the aggregated radar data (network input). Table entries are intersection-over-union for each of the classes and average over all 3 classes (last column). Both ray trace and occupancy net are with 20 frame aggregation. It is clear our occupancy net outperforms the classic approaches as well as the ray trace baseline in all categories by a large margin.}
    \label{tab:results}
\end{table}

\newcommand{\ic}[1]{\includegraphics[width=1.4cm, height=4.9cm ]{#1}}
\begin{figure*}[t]
    \centering
    \begin{tabular}{c c c c c c c c c}
    \ic{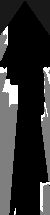} & 
    \ic{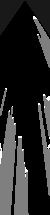} &
    \ic{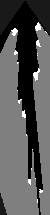} & 
    \ic{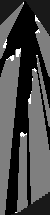} & 
    \ic{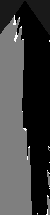} & 
    \ic{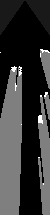} & 
    \ic{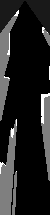} & 
    \ic{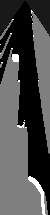} & 
    \ic{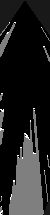} \\
    \ic{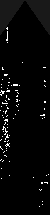} & 
    \ic{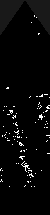} & 
    \ic{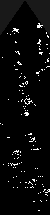} &
    \ic{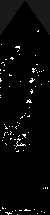} & 
    \ic{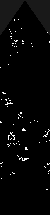} & 
    \ic{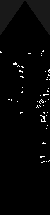} &
    \ic{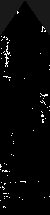} & 
    \ic{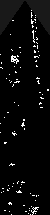} & 
    \ic{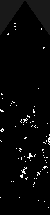} \\
    \ic{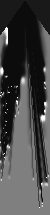} & 
    \ic{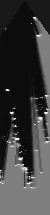} & 
    \ic{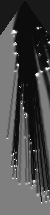} & 
    \ic{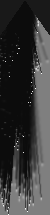} & 
    \ic{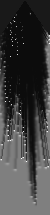} & 
    \ic{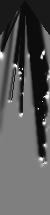} & 
    \ic{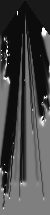} & 
    \ic{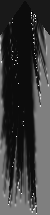} & 
    \ic{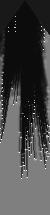} \\
    \ic{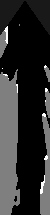} & 
    \ic{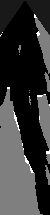} & 
    \ic{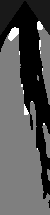} & 
    \ic{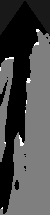} & 
    \ic{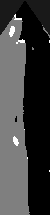} & 
    \ic{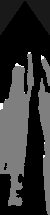} & 
    \ic{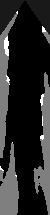} & 
    \ic{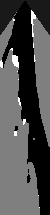} & 
    \ic{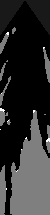} \\
    (a) & (b) & (c) & (d) & (e) & (f) & (g) & (h) & (i) \\
    \end{tabular}
    \caption{Qualitative results from 3 different radar positions (front, back left, back right). Columns show different examples. Top row showing ground truth, second row radar input, third row classic ISM and bottom row our occupancy net predictions with 20 frame aggregation. Network is agnostic to radar position therefore the same network model is used from all radars. The radar data is very noisy making this a challenging task. It is evident the learned grid maps are much sharper and more accurate than the ones produced by the classic method. Note: White is occupied, black is free, light gray is unobserved, dark gray is ignore.}
    \label{fig:results}
\end{figure*}
\section{Conclusions}\label{sec:conclusions}
Radar is an emerging sensor in autonomous vehicle vision that can be leveraged for road scene understanding in challenging scenarios and conditions. 

In this work we focused on occupancy grid creation from radar data. This task was made especially challenging as we were  dealing with clustered radar data (and not raw). Despite recent advances in computer vision and deep learning technology, in the vast majority of studies, occupancy grid mapping from radar data is still performed using classic filtering techniques.  
In this work we have shown that learning occupancy grid mapping from clustered radar data is feasible and represents an attractive solution which significantly outperforms commonly used standard techniques. We have formulated the problem as a computer vision task of learning a three class semantic segmentation problem. This allows understanding which space is free and which is occupied, and also explicitly infer which parts of the space are unobserved. We have shown how lidar data can be used to generate training labels.
We leveraged temporal data to handle data sparsity, and addressed class imbalance.

We believe this work and results achieved using the NuScenes data set provide a baseline for this fundamental task and help promoting further research. 
Possible future directions of this work are handling dynamic objects, adding an explicit spatial consistency mechanism, and exploring additional techniques for handling temporal information. In addition, our formulation serves as a basis for sensor fusion, which is a prominent research direction.

\clearpage
{\small
\bibliographystyle{ieee}
\bibliography{bib}

\begin{thebibliography}{10}\itemsep=-1pt

\bibitem{badrinarayanan2017segnet}
V.~Badrinarayanan, A.~Kendall, and R.~Cipolla.
\newblock Segnet: A deep convolutional encoder-decoder architecture for image
  segmentation.
\newblock {\em IEEE transactions on pattern analysis and machine intelligence},
  39(12):2481--2495, 2017.

\bibitem{berman2018}
M.~{Berman}, A.~R. {Triki}, and M.~B. {Blaschko}.
\newblock The lovasz-softmax loss: A tractable surrogate for the optimization
  of the intersection-over-union measure in neural networks.
\newblock In {\em 2018 IEEE/CVF Conference on Computer Vision and Pattern
  Recognition}, pages 4413--4421, June 2018.

\bibitem{caesar2019nuscenes}
H.~Caesar, V.~Bankiti, A.~H. Lang, S.~Vora, V.~E. Liong, Q.~Xu, A.~Krishnan,
  Y.~Pan, G.~Baldan, and O.~Beijbom.
\newblock nuscenes: A multimodal dataset for autonomous driving.
\newblock {\em arXiv preprint arXiv:1903.11027}, 2019.

\bibitem{chavez2016multiple}
R.~O. Chavez-Garcia and O.~Aycard.
\newblock Multiple sensor fusion and classification for moving object detection
  and tracking.
\newblock {\em IEEE Transactions on Intelligent Transportation Systems},
  17(2):525--534, 2016.

\bibitem{degerman20163d}
J.~Degerman, T.~Pernst{\aa}l, and K.~Alenljung.
\newblock 3d occupancy grid mapping using statistical radar models.
\newblock In {\em 2016 IEEE Intelligent Vehicles Symposium (IV)}, pages
  902--908. IEEE, 2016.

\bibitem{dickmann2016automotive}
J.~Dickmann, J.~Klappstein, M.~Hahn, N.~Appenrodt, H.-L. Bloecher, K.~Werber,
  and A.~Sailer.
\newblock Automotive radar the key technology for autonomous driving: From
  detection and ranging to environmental understanding.
\newblock In {\em 2016 IEEE Radar Conference (RadarConf)}, pages 1--6. IEEE,
  2016.

\bibitem{edelsbrunner1983shape}
H.~Edelsbrunner, D.~Kirkpatrick, and R.~Seidel.
\newblock On the shape of a set of points in the plane.
\newblock {\em IEEE Transactions on information theory}, 29(4):551--559, 1983.

\bibitem{elfes1989using}
A.~Elfes.
\newblock Using occupancy grids for mobile robot perception and navigation.
\newblock {\em Computer}, 22(6):46--57, 1989.

\bibitem{garcia2008high}
R.~Garcia, O.~Aycard, T.-D. Vu, and M.~Ahrholdt.
\newblock High level sensor data fusion for automotive applications using
  occupancy grids.
\newblock In {\em 2008 10th International Conference on Control, Automation,
  Robotics and Vision}, pages 530--535. IEEE, 2008.

\bibitem{gohring2011radar}
D.~G{\"o}hring, M.~Wang, M.~Schn{\"u}rmacher, and T.~Ganjineh.
\newblock Radar/lidar sensor fusion for car-following on highways.
\newblock In {\em The 5th International Conference on Automation, Robotics and
  Applications}, pages 407--412. IEEE, 2011.

\bibitem{hoermann2018dynamic}
S.~Hoermann, M.~Bach, and K.~Dietmayer.
\newblock Dynamic occupancy grid prediction for urban autonomous driving: A
  deep learning approach with fully automatic labeling.
\newblock In {\em 2018 IEEE International Conference on Robotics and Automation
  (ICRA)}, pages 2056--2063. IEEE, 2018.

\bibitem{kellner2012grid}
D.~Kellner, J.~Klappstein, and K.~Dietmayer.
\newblock Grid-based dbscan for clustering extended objects in radar data.
\newblock In {\em 2012 IEEE Intelligent Vehicles Symposium}, pages 365--370.
  IEEE, 2012.

\bibitem{li2018high}
M.~Li, Z.~Feng, M.~Stolz, M.~Kunert, R.~Henze, and F.~K{\"u}{\c{c}}{\"u}kay.
\newblock High resolution radar-based occupancy grid mapping and free space
  detection.
\newblock In {\em VEHITS}, pages 70--81, 2018.

\bibitem{lin2017feature}
T.-Y. Lin, P.~Doll{\'a}r, R.~Girshick, K.~He, B.~Hariharan, and S.~Belongie.
\newblock Feature pyramid networks for object detection.
\newblock In {\em Proceedings of the IEEE Conference on Computer Vision and
  Pattern Recognition}, pages 2117--2125, 2017.

\bibitem{lombacher2017semantic}
J.~Lombacher, K.~Laudt, M.~Hahn, J.~Dickmann, and C.~W{\"o}hler.
\newblock Semantic radar grids.
\newblock In {\em 2017 IEEE Intelligent Vehicles Symposium (IV)}, pages
  1170--1175. IEEE, 2017.

\bibitem{lu2019monocular}
C.~Lu, M.~J.~G. van~de Molengraft, and G.~Dubbelman.
\newblock Monocular semantic occupancy grid mapping with convolutional
  variational encoder--decoder networks.
\newblock {\em IEEE Robotics and Automation Letters}, 4(2):445--452, 2019.

\bibitem{oh2016fast}
S.-I. Oh and H.-B. Kang.
\newblock Fast occupancy grid filtering using grid cell clusters from lidar and
  stereo vision sensor data.
\newblock {\em IEEE Sensors Journal}, 16(19):7258--7266, 2016.

\bibitem{prophet2018adaptions}
R.~Prophet, H.~Stark, M.~Hoffmann, C.~Sturm, and M.~Vossiek.
\newblock Adaptions for automotive radar based occupancy gridmaps.
\newblock In {\em 2018 IEEE MTT-S International Conference on Microwaves for
  Intelligent Mobility (ICMIM)}, pages 1--4. IEEE, 2018.

\bibitem{qian1999momentum}
N.~Qian.
\newblock On the momentum term in gradient descent learning algorithms.
\newblock {\em Neural networks}, 12(1):145--151, 1999.

\bibitem{rexin2017modeling}
N.~Rexin, D.~Nuss, S.~Reuter, and K.~Dietmayer.
\newblock Modeling occluded areas in dynamic grid maps.
\newblock In {\em 2017 20th International Conference on Information Fusion
  (Fusion)}, pages 1--6. IEEE, 2017.

\bibitem{ronneberger2015u}
O.~Ronneberger, P.~Fischer, and T.~Brox.
\newblock U-net: Convolutional networks for biomedical image segmentation.
\newblock In {\em International Conference on Medical image computing and
  computer-assisted intervention}, pages 234--241. Springer, 2015.

\bibitem{skolnik2008radar}
M.~I. Skolnik.
\newblock Radar handbook 3rd edition.
\newblock 2008.

\bibitem{thrun2005probabilistic}
S.~Thrun, W.~Burgard, and D.~Fox.
\newblock {\em Probabilistic robotics}.
\newblock 2005.

\bibitem{weiss2007robust}
T.~Weiss, B.~Schiele, and K.~Dietmayer.
\newblock Robust driving path detection in urban and highway scenarios using a
  laser scanner and online occupancy grids.
\newblock In {\em 2007 IEEE Intelligent Vehicles Symposium}, pages 184--189.
  IEEE, 2007.

\bibitem{werber2015automotive}
K.~Werber, M.~Rapp, J.~Klappstein, M.~Hahn, J.~Dickmann, K.~Dietmayer, and
  C.~Waldschmidt.
\newblock Automotive radar gridmap representations.
\newblock In {\em 2015 IEEE MTT-S International Conference on Microwaves for
  Intelligent Mobility (ICMIM)}, pages 1--4. IEEE, 2015.

\bibitem{weston2018probably}
R.~Weston, S.~Cen, P.~Newman, and I.~Posner.
\newblock Probably unknown: Deep inverse sensor modelling in radar.
\newblock {\em arXiv preprint arXiv:1810.08151}, 2018.

\end{thebibliography}
}

\end{document}